\documentclass[10pt,letterpaper,twocolumn]{article}
\usepackage{array}
\usepackage{multirow}
\usepackage{amsmath}
\usepackage{amssymb}
\usepackage{graphicx}
\usepackage{float}
\usepackage{epstopdf}
\usepackage{url}

\usepackage{booktabs}

\makeatletter

\pdfpageheight\paperheight
\pdfpagewidth\paperwidth

\usepackage{cvpr}
\usepackage{times}
\usepackage{epsfig}

\usepackage{algorithmicx}
\usepackage[ruled]{algorithm}
\usepackage[noend]{algpseudocode}
\usepackage{setspace}

\cvprfinalcopy 

\def\cvprPaperID{949} 

\makeatother

\begin{document}

\title{Sketch-based 3D Shape Retrieval using Convolutional Neural Networks}

\author{Fang Wang$^{1}$ $~~~~~~~~~$ $~~~~~~~~~$Le Kang$^{2}$
$~~~~~~~~~$ $~~~~~~~~~$Yi Li$^{1}$\\
\\
 $^{1}$NICTA and ANU$~~~~~~~~~$ $~~~~~~~~~$ $^2$ECE, University of Maryland at College Park\\
$^{1}$\{fang.wang, yi.li\}@nicta.com.au, $^{2}$lekang@umiacs.umd.edu
}

\maketitle
\begin{abstract}
Retrieving 3D models from 2D human sketches has received considerable attention in the areas of graphics, image retrieval, and computer vision.
Almost always in state of the art approaches a large amount of ``best views'' are computed for 3D models, with the hope that the query sketch matches one of these 2D projections of 3D models using predefined features.

We argue that this two stage approach (view selection -- matching) is pragmatic but also problematic because the ``best views'' are subjective and ambiguous, which makes the matching inputs obscure. This imprecise nature of matching further makes it challenging to choose features manually.
Instead of relying on the elusive concept of ``best views'' and the hand-crafted features, we propose to define our views using a minimalism approach and learn features for both sketches and views.
Specifically, we drastically reduce the number of views to only two predefined directions for the whole dataset. 
Then, we learn two Siamese Convolutional Neural Networks (CNNs), one for the views and one for the sketches.
The loss function is defined on the within-domain as well as the cross-domain similarities.
Our experiments on three benchmark datasets demonstrate that our method is significantly better than state of the art approaches, and outperforms them in all conventional metrics.
\end{abstract}

\section{Introduction}

Retrieving 3D models from 2D sketches has important applications in computer graphics, information retrieval, and computer vision \cite{eitz2012sbsr, furuya2013ranking, li2014comparison}.
Compared to the early attempts where keywords or 3D shapes are used as queries \cite{shilane2004princeton}, the sketch-based idea is very attractive because sketches by hand provide an easy way to input, yet they are rich enough to specify shapes. 

\begin{figure}[t]
\begin{center}
\begin{tabular}{c}
	\includegraphics[width=0.75\columnwidth]{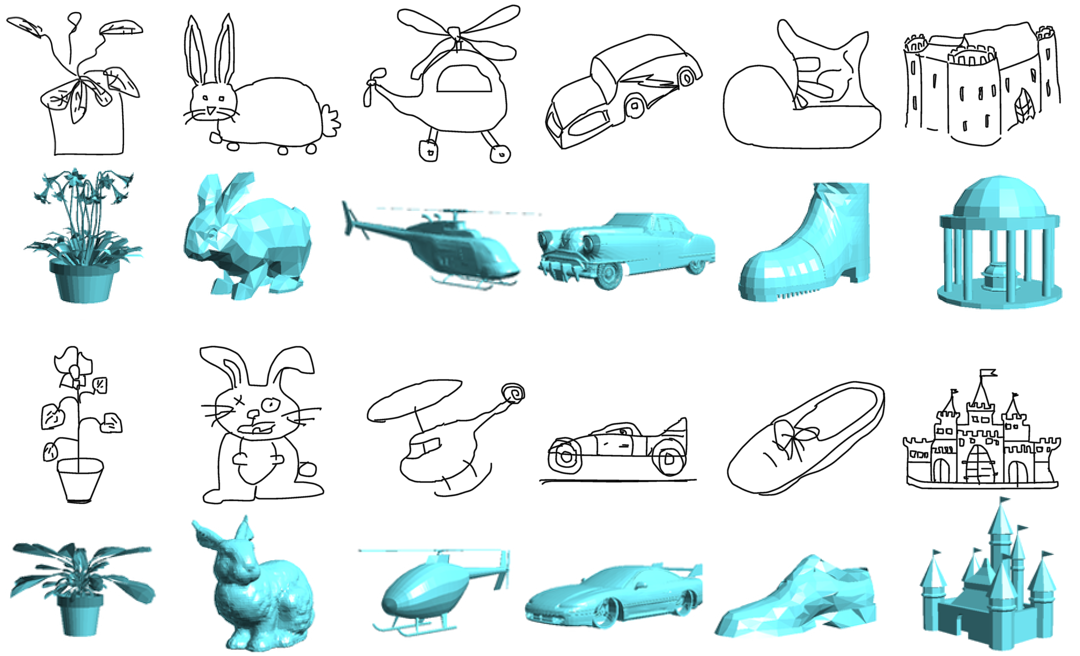} \\
\end{tabular}
\end{center}
\caption{Examples of sketch based 3D shape retrieval.\label{fig:examples}}
\end{figure}

Directly matching 2D sketches to 3D models suffers from significant differences between the 2D and 3D representations.
Thus, in many state of the art methods 3D models are projected to multiple 2D views, and a sketch matches a 3D model if it matches one of its views.
Fig. \ref{fig:examples} shows a few examples of 2D sketches and their corresponding 3D models.
One can immediately see the variations in both the sketch styles and 3D models.

In almost all state of the art approaches, sketch based 3D shape retrieval amounts to finding the ``best views'' for 3D models and hand-crafting the right features for matching sketches and views.
First, an automatic procedure is used to select the most representative views of a 3D model. 
Ideally, one of the viewpoints is similar to that of the query sketches.
Then, 3D models are projected to 2D planes using a variety of line rendering algorithms.
Subsequently, many 2D matching methods can be used for computing the similarity scores, where features are always manually defined (\textit{e.g.}, Gabor, dense SIFT, and GALIF \cite{eitz2012sbsr}).

This stage-wise methodology appears pragmatic, but it also brings a number of puzzling issues.
To begin with, there is no guarantee that the best views have similar viewpoints with the sketches.
The inherent issue is that identifying the best views is an unsolved problem on its own, partially because the general definition of best views is elusive.
In fact, many best view methods require manually selected viewpoints for training, which makes the view selection by finding ``best views'' a chicken-egg problem.

Further, this viewpoint uncertainty makes it dubious to match samples from two different domains without learning their metrics.
Take Fig. \ref{fig:examples} for example, even when the viewpoints are similar the variations in sketches as well as the different characteristics between sketches and views are beyond the assumptions of many 2D matching methods.

Considering all the above issues arise when we struggle to seek the viewpoints for matching, can we bypass the stage of view selection? 
In this paper we demonstrate that by learning cross domain similarities, we no longer require the seemingly indispensable view similarity assumption.

Instead of relying on the elusive concept of ``best views'' and hand-crafted features, we propose to define our views and learn features for views and sketches.
Assuming that the majority of the models are upright, we drastically reduce the number of views to two per object for the whole dataset.
We also make no selections of these two directions as long as they are significantly different.
Therefore, we consider this as the minimalism approach as opposed to multiple best views.

This upright assumption appears to be strong, but it turns out to be sensible for 3D datasets.
Many 3D models are naturally generated upright (\textit{e.g.}, \cite{shilane2004princeton}).
We choose two viewpoints because it is very unlikely to get degenerated views for two significantly different viewpoints.
An immediate advantage is that our matching is more efficient without the need of comparing to more views than necessary.

This seemingly radical approach triumphs only when the features are learned properly.
In principle, this can be regarded as learning representations between sketches and views by specifying similarities, which gives us a semantic level matching. 
To achieve this, we need comprehensive shape representations rather than the combination of  shallow features that only capture low level visual information. 

We learn the shape representations using Convolutional Neural Network (CNN). 
Our model is based on the Siamese network \cite{chopra2005learning}.
Since the two input sources have distinctive intrinsic properties, we use two different CNN models, one for handling the sketches and the other for the views.
This two model strategy can give us more power to capture different properties in different domains.

Most importantly, we define a loss function to ``align'' the results of the two CNN models.
This loss function couples the two input sources into the same target space, which allows us to compare the features directly using a simple distance function. 

Our experiments on three large datasets show that our method significantly outperforms state of the art approaches in a number of metrics, including precision-recall and the nearest neighbor.
We further demonstrate the retrievals in each domain are effective.
Since our network is based on filtering, the computation is fast.

Our contributions include
\begin{itemize}
\item We propose to learn feature representations for sketch based shape retrieval, which bypasses the dilemma of best view selection;
\item We adopt two Siamese Convolutional Neural Networks to successfully learn similarities in both the within-domain and the cross domain;
\item We outperform all the state of the art methods on three large datasets significantly.
\end{itemize}

\section{Related work}

Sketch based shape retrieval has received many interests for years \cite{Funkhouser:2003:ASE}.
In this section we review three key components in sketch based shape retrieval: public available datasets, features, and similarity learning.

\vspace{-5pt}
\paragraph*{Datasets}
The effort of building 3D datasets can be traced back to decades ago. 
The Princeton Shape Benchmark (PSB) is probably one of the best known sources for 3D models \cite{shilane2004princeton}. 
There are some recent advancements for general and special objects, such as the SHREC'14 Benchmark \cite{LLL+14a} and the Bonn Architecture Benchmark \cite{Wessel:2009:SBR:2381128.2381141}. 

2D sketches have been adopted as input in many systems \cite{journals/ijcv/DarasA10}. 
However, the large scale collections are available only recently.
Eitz \textit{et al.} \cite{eitz2012sbsr} collected sketches based on the PSB dataset. Li \textit{et al.} \cite{li2014comparison} organized the sketches collected by \cite{eitz2012hdhso} in their SBSR challenge.

\vspace{-5pt}
\paragraph*{Features}
Global shape descriptors, such as statistics of shapes \cite{Osada:2002:SD} and distance functions \cite{Kazhdan:2002:RSD:645316.649205}, have been used for 3D shape retrieval \cite{Tangelder:2008:SCB:1395016.1395041}. 
Recently, local features is proposed for partial matching \cite{Funkhouser:2006:PMS:1281957.1281974} or used in the bag-of-words model for 3D shape retrieval \cite{Bronstein:2011:SGG:1899404.1899405}.

Boundary information together with internal structures are used for matching sketches against 2D projections. 
Therefore, a good representation of line drawing images is a key component for sketch based shape retrieval.
Sketch representation such as shape context \cite{belongie2002shape} was proposed for image based shape retrieval.
Furuya \textit{et al.} proposed BF-DSIFT feature, which is an extended SIFT feature with Bag-of-word method, to represent sketch images \cite{furuya2009dense}.
One recent method is the Gabor local line based feature (GALIF) by Mathias \textit{et al.}, which builds on a bank of Gabor filters followed by a Bag-of-word method \cite{eitz2012sbsr}. 

In addition to 2D shape features, some methods also explored geometry features as well as graph-based features to facilitate the 3D shape retrieval \cite{li2015comparison}. 
Semantic labeling is also used to bridge the gaps between different domains \cite{gong2013learning}.
In this paper, we focus on view based method and only use 2D shape features.

\vspace{-5pt}
\paragraph*{CNN and Siamese network}
Recently deep learning has achieved great success on many computer vision tasks. Specifically, CNN has set records on standard object recognition benchmarks \cite{Krizhevsky2012}. With a deep structure, the CNN can effectively learn complicated mappings from raw images to the target, which requires less domain knowledge compared to handcrafted features and shallow learning frameworks.

A Siamese network \cite{chopra2005learning} is a particular neural network architecture consisting of two identical sub-convolutional networks, which is used in a weakly supervised metric learning setting.
The goal of the network is to make the output vectors similar if input pairs are labeled as similar, and dissimilar for the input pairs that are labeled as dissimilar.
Recently, the Siamese network has been applied to text classification \cite{Yih:2011:LDP:2018936.2018965} and speech feature classification \cite{NIPS2011_4314}.

\section{Learning feature representations for sketch based 3D shape retrieval}

We first briefly introduce basic concepts in CNNs and Siamese network.
Then, we present our network architecture for cross domain matching, based on the Siamese network.
Given a set of view and sketch pairs, we propose to use two different Siamese networks, one for each domain.
Finally, we revisit the view selection problem, and describe our minimalism approach of viewpoint definition and the line drawing rendering procedure.

\subsection{CNN and Siamese network}

CNN is a multilayer learning framework, which consists of an input layer, a few convolutional layers and fully connected layers, as well as an output layer on which the loss function is defined. The goal of CNN is to learn a hierarchy of feature representations. Signals in each layer are convolved with a number of filters and further downsampled by pooling operations, which aggregate values in a small region by functions including max, min, and average. 
The learning of CNN is based on Stochastic Gradient Descent (SGD). Please refer to \cite{LeCun:1998:CNI:303568.303704} for details.

Siamese Convolutional Neural Network has been used successfully for dimension reduction in weakly supervised metric learning.
Instead of taking single sample as input, the network typically takes a pair of samples, and the loss functions are usually defined over pairs.
A typical loss function of a pair has the following form:
\begin{align}
L(s_1, s_2, y) &= (1-y) \alpha D_w^2 + y \beta e^{\gamma D_w},
\label{eq:loss}
\end{align}
where $s_1$ and $s_2$ are two samples, $y$ is the binary similarity label, $D_w = {\Vert f(s_1; w_1)-f(s_2; w_2) \Vert}_1$ is the distance.
Following \cite{chopra2005learning}, we set $\alpha = \frac{1}{C_p}$, $\beta = C_n$, and $\gamma = \frac{-2.77}{C_n}$, where $C_p=0.2$ and $C_n=10$ are two constants.

This can be regarded as a metric learning approach.
Unlike methods that assign binary similarity labels to pairs, the network aims at bring the output feature vectors closer for input pairs that are labeled as similar, or push the feature vectors away if the input pairs are labeled as dissimilar.

The Siamese network is frequently illustrated as two identical networks for two different samples.
In each SGD iteration, pairs of samples are processed using two identical networks, and the error computed by Eq. \ref{eq:loss} is then back-propagated and the gradients are computed individually base on the two sample sets.
The Siamese network is updated by the average of these two gradients.

\subsection{Cross-domain matching using Siamese network}

In this section, we propose a method to match samples from two domains without the heavy assumption of view similarity.
We first provide our motivation using an illustrated sample.
Then, we propose our extension of the basic Siamese network.
Specifically, we use two different networks to handle sources from different domains.

\subsubsection{An illustrated example}

The matching problem in sketch based shape retrieval can be seen as a metric learning paradigm.
In each domain, the samples are mapped to some feature vectors.
The cross domain matching is successful if the features from each domain are ``aligned'' correctly.
\begin{figure}[h]
\begin{center}
\begin{tabular}{c|c}
	\includegraphics[width=0.34\columnwidth]{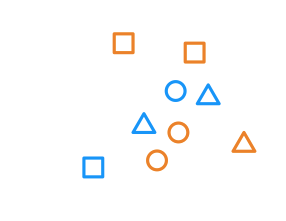} & 
	\includegraphics[width=0.34  \columnwidth]{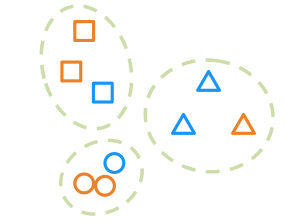} \\
	(a) & (b)
\end{tabular}
\end{center}
\caption{An illustrated example, a) the shapes in the original domain may be mixed, and b) after cross-domain metric learning, similar shapes in both domains are grouped together.}
\label{fig:illustration}
\end{figure}

This idea is illustrated in Fig. \ref{fig:illustration}.
Blue denotes samples in the sketch domain, and the orange denotes the ones in the view domain.
Different shapes denote different classes.
Before learning, the feature points from two different domains are initially mixed together
(Fig. \ref{fig:illustration}a).
If we learn the correct mapping using pair similarities in each domain as well as their cross-domain relations jointly, the two point sets may be correctly aligned in the feature space (Fig. \ref{fig:illustration}b). 
After this cross domain metric learning, matching can be performed in both the same domain (sketch-sketch and view-view) and cross domain (sketch-view). 

Note that, there are no explicit requirements about viewpoint similarity in this perspective ({\it i.e.,} whether the matched pairs are from the same viewpoints is less important).
Instead, the focus is the metric between the two domains and the mapping within the same domain.

\subsubsection{Two networks, one loss}
\begin{figure*}[t]
\begin{center}
\includegraphics[width=0.75\textwidth]{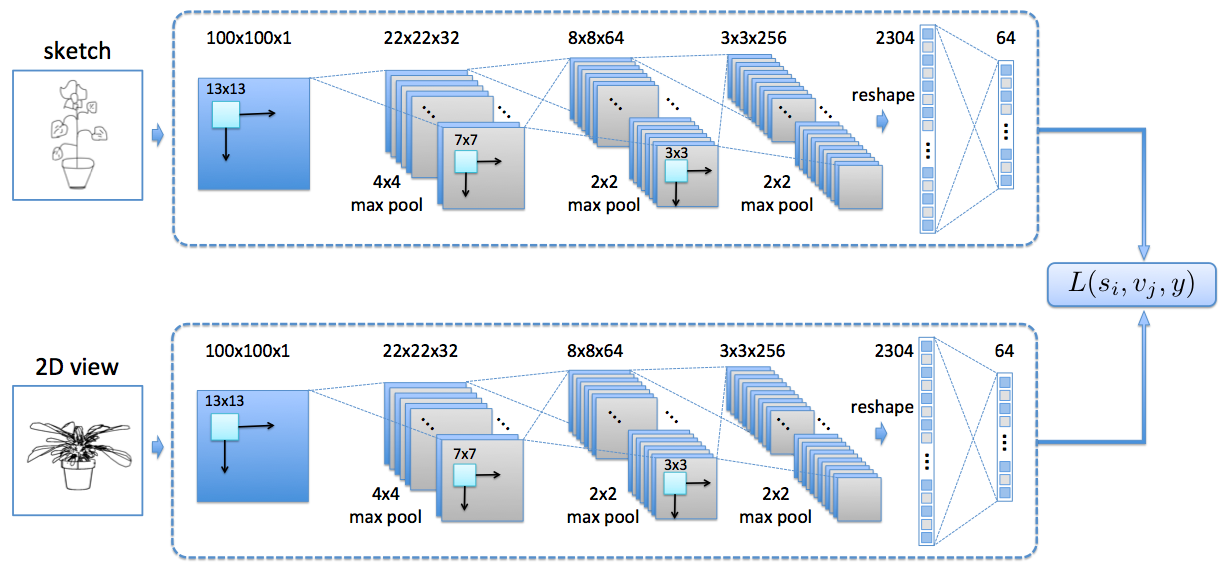} 
\end{center}

\caption{Dimension reduction using Siamese network.\label{fig:Siamese}}
\end{figure*}

The basic Siamese network is commonly used for samples from the same domain. In the cross domain setting, we propose to extend the basic version to two Siamese networks, one for the view domain and the other for the sketch domain.
Then, we define the within-domain loss and the cross domain loss.
This hypothesis is supported in the Sec. \ref{sec:exp}.

Assuming we have two inputs from each domain, \textit{i.e.}, $s_1$ and $s_2$ are two sketches and $v_1$ and $v_2$ are two views. 
For simplicity, we assume $s_1$ and $v_1$ are from the same class and $s_2$ and $v_2$ are from the same class as well.
Therefore, one label $y$ is enough to specify their relationships.

As a result, our loss function is composed by three terms: the similarity of sketches, the similarity of views, and the cross domain similarity. 
\begin{align}
\mathcal{L}(s_1,& s_2, v_1, v_2, y) \nonumber
\\ 
                         &= L(s_1,s_2,y) + L(v_1,v_2,y) + L(s_1,v_1,y),
\label{eq:finalloss}
\end{align}
where $L(\cdot, \cdot, \cdot)$ is defined by Eq. \ref{eq:loss}.
Please note that, while the category information available in the dataset can be exploited to improve the performance, we do not use the category labels in the above framework. 

\subsection{Network architecture}

Fig. \ref{fig:Siamese} shows the architecture of our network for the inputs being views and sketches, respectively.

We use the same network design for both networks, but they are learned separately.
Our input patch size is $100 \times 100$ for both sources.
The structure of the single CNN has three convolutional layers, each with a max pooling, one fully connected layer to generate the features, and one output layer to compute the loss (Eq. \ref{eq:finalloss}). 

The first convolutional layer followed by a $4\times4$ pooling generates 32 response maps, each of size $22\times22$. 
The second layer and pooling outputs 64 maps of size $8\times8$.
The third layer layer has 256 response maps, each pooled to a size of $3\times3$. 
The 2304 features generated by the final pooling operation are linearly transformed to $64 \times 1$ features in the last layer.
Rectified linear units are used in all layers. 

\subsection{View definitions and line drawing rendering}

We present our procedure of generating viewpoints and rendering 3D models.
As opposed to multiple views, we find it sufficient to use two views to characterize a 3D model because the chance that both views are degenerated is little.
Following this observation, we impose the minimal assumptions on choosing views for the whole dataset:
\begin{enumerate}
  \item Most of the 3D models in the dataset are up-right;
  \item Two viewpoints are randomly generated for the whole dataset, provided that the difference in their angles is larger than $45$ degrees.
\end{enumerate}

\begin{figure}[h]
\begin{center}
\begin{tabular}{c}
	\includegraphics[width=0.65\columnwidth]{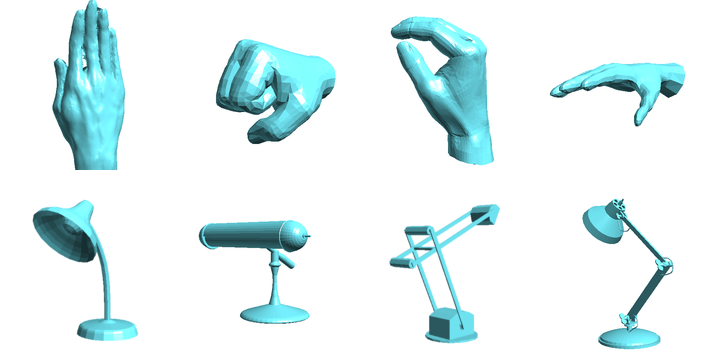}\\
\end{tabular}
\end{center}
\caption{3D models viewed from predefined viewpoints.}
\label{fig:views}
\end{figure}

Fig. \ref{fig:views} shows some of our views in the PSB dataset.
The first row shows that the upright assumption does not require strict alignments of 3D models, because some models may not have well defined orientation.
Further, while the models are upright, they can still has different rotations. 

We want to stress that our approach does not eliminate the possibility of selecting more (best) views as input, but the comparisons among view selection methods are beyond the scope of this paper.

\begin{figure}[h]
\begin{center}
\begin{tabular}{ccc}
	\includegraphics[width=0.2\columnwidth]{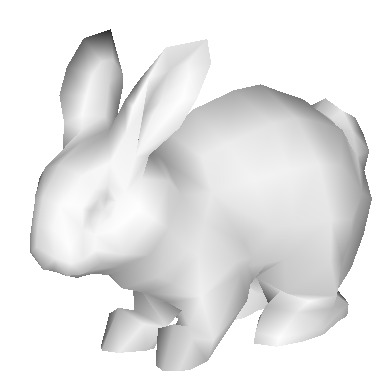} &
	\includegraphics[width=0.2\columnwidth]{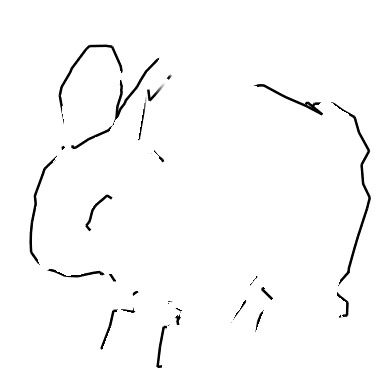} &
	\includegraphics[width=0.2\columnwidth]{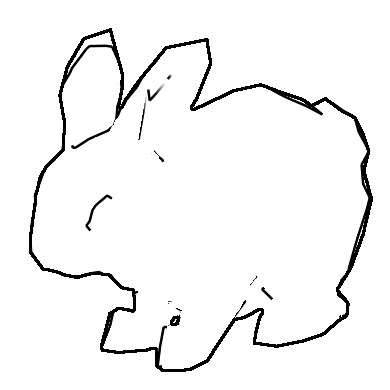} \\
	(a) Shaded & (b) SC & (c) Final
\end{tabular}
\end{center}
\caption{Rendering 3D models.}

\label{fig:rendering}
\end{figure}

Once the viewpoints are chosen, we render the 3D models and generate 2D line drawings.
Rendering line drawings that include strong abstraction and stylization effects is a very useful topic in computer graphics, computer vision, and psychology.
Outer edges and internal edges both play an important role in this rendering process.
Therefore, we use the following descriptors: 1) closed boundaries and 2) Suggestive Contours \cite{DeCarlo:2003:SCF} (Fig. \ref{fig:rendering}).

\section{Experiments}\label{sec:exp}

We present our experiments on three recent large datasets in this section.
In all experiments our method outperforms the state of the arts in a number of well recognized metrics.
In additional to the cross-domain retrieval, we also present our within-domain retrieval results, which have not been reported in any other comparison methods.
These experiments demonstrate that our Siamese network successfully learns the feature representations for both domains.
The data and the code is available at \url{http://users.cecs.anu.edu.au/~yili/cnnsbsr/}.

\subsection{Datasets}

\paragraph{PSB / SBSR dataset}
The Princeton Shape Benchmark (PSB) \cite{shilane2004princeton} is widely used for 3D shape retrieval system evaluation, which contains 1814 3D models and is equally divided into training set and testing set.

In \cite{eitz2012sbsr}, the Shape Based Shape Retrieval (SBSR) dataset is collected based on the PSB dataset. 
The 1814 hand drawn sketches are collected using Amazon Mechanical Turk. 
In the collection process, participants are asked to draw sketches given only the name of the categories without any visual clue from the 3D models. 

\paragraph{SHREC'13 \& '14 dataset}
Although the PSB dataset is widely used in shape retrieval evaluation, there is a concern that the number of sketches for each class in the SBSR dataset is not enough. Some classes have only very few instances (27 of 90 training classes have no more than 5 instances), while some classes have dominating number of instances, \textit{e.g.}, the ``fighter jet" class and the ``human" class have as many as 50 instances. 

To remove the possible bias when evaluating the retrieval algorithms, Li \textit{et al.} \cite{li2014comparison} reorganized the PSB/SBSR dataset, and proposed a SHREC'13 dataset where a subset of PSB with 1258 models is used and the sketches in each classes has 80 instances.
These sketch instances are split in two sets: 50 for training and 30 for testing. 
Please note, the number of models in each class still varies.
For example, the largest class has 184 instances but there are 23 classes containing no more than 5 models

Recently, SHREC'14 is proposed to address some above concerns \cite{LLL+14a}, which greatly enlarges the number of 3D models to 8987, and the number of classes is doubled. The large variation of this dataset makes it much more challenging, and the overall performance of all reported methods are very low (\textit{e.g.}, the accuracy for the best algorithm is only 0.16 for the top 1 candidate). 
This is probably due to the fact that the models are from various sources and are arbitrarily oriented.
While our performance is still superior (see Fig. \ref{fig:SHRECstat}b and Table. \ref{tab:shreccmp}), we choose to present our results using the SHREC'13 dataset.


\begin{figure*}[htbp]
\begin{center}
\begin{tabular}{c}
	\includegraphics[width=0.85\textwidth]{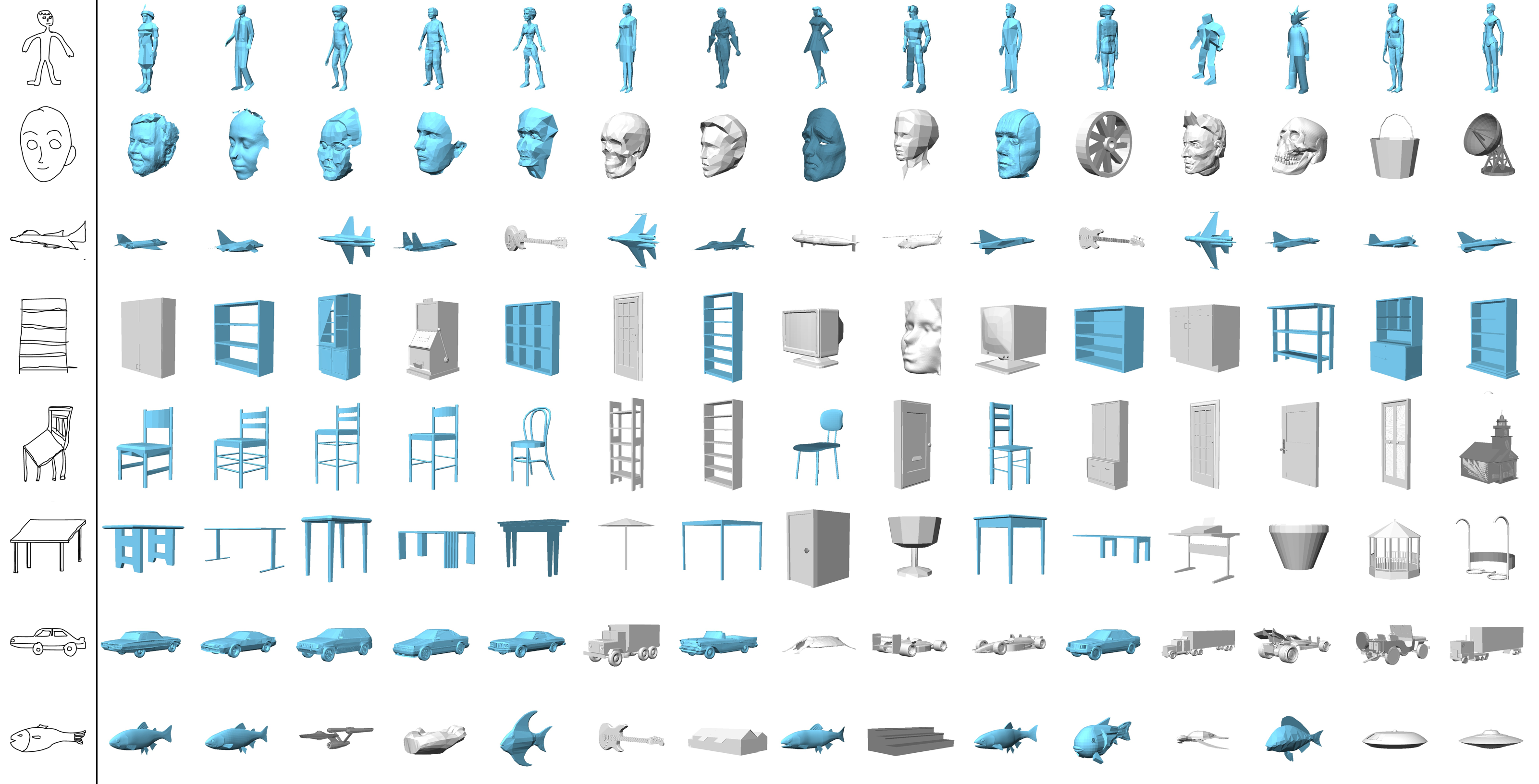}
\end{tabular}
\end{center}
\caption{Retrieval examples of PSB/SBSR dataset. Cyan denotes the correct retrievals.}
\label{fig:tub}
\end{figure*}

\vspace{-10pt}
\paragraph{Evaluation criteria}
In our experiment, we use the above datasets and measure the performance using the following criteria:
1) \textit{Precision-recall curve} is calculated for each query and linear interpolated, then the final curve is reported by averaging all precision values for fixed recall rates;
2) \textit{Average precision (mAP)} is the area under the precision-recall curve;
3) \textit{Nearest neighbor (NN)} is used to measure the top 1 retrieval accuracy; 
4) \textit{E-Measure (E)} is the harmonic mean of the precision and recall for the top 32 retrieval results;
5) \textit{First/second tier (FT/ST)} and \textit{Discounted cumulated gain (DCG)} as defined in the PSB statistics. 

\subsection{Experimental settings}

\paragraph{Stopping criteria}
All three of the datasets had been split into training and testing sets, but no validation set was specified. Therefore, we terminated our algorithm after $50$ epochs for PSB/SBSR and $20$ for SHREC'13 dataset (or until convergence).
Multiple runs were performed and the mean values were reported.

\vspace{-5pt}
\paragraph{Generating pairs for Siamese network}
To make sure we generate reasonable proportion of similar and dissimilar pairs, we use the following approach to generate pair sets.
For each training sketch, we random select $k_p$ view pairs in the same category (matched pairs) and $k_n$ view samples from other categories (unmatched pairs).
Usually, our dissimilar pairs are ten times more than the similar pairs for successful training.
In our experiment, we use $k_p=2$, $k_n=20$. 
We perform this random pairing for each training epoch. 
To increase the number of training samples, we also used data augmentation for the sketch set. 
To be specific, we randomly perform affine transformations on each sketch sample with small scales and angles to generate more variations. We generate two augmentations for each sketch sample in the dataset. 



\paragraph{Computational cost}
The implementation of the proposed Siamese CNN is based on the Theano \cite{Theano} library.
We measure the processing time on on a PC with 2.8GHz CPU and GTX 780 GPU.
With preprocessed view features, the retrieval time for each query is approximately 0.002 sec on average on SHREC'13 dataset.

The training time is proportional to the total number of pairs and the number of epochs.
Overall training  takes approximately 2.5 hours for PSB/SBSR, 6 hours for SHREC'13, respectively.
Considering the total number of pairs is large, the training time is sensible.

We test various number of views in our experiments. We find that there was no significant performance gain when we vary the view from two to ten. However, it increased the computational cost significantly when more views are used, and more importantly, the GPU memory. This motivates us to select only two views in the experiments.

\subsection{Shape retrieval on PSB/SBSR dataset}

\subsubsection{Examples}

In this section, we test our method using the PSB/SBSR dataset.
First, we show some retrieval examples in Fig. \ref{fig:tub}.
The first column shows 8 queries from different classes, and each row shows the top 15 retrieval results.
Cyan denotes the correct retrievals, and gray denotes incorrect ones.

Our method performs exceptionally well in popular classes such as human, face, and plane.
We also find that some fine grained categorizations are difficult to distinguish.
For instance, the shelf and the box differ only in a small part of the model.
However, we also want to note that some of the classes only differ in semantics (\textit{e.g.}, barn and house only differ in function).
Certainly, this semantic ambiguity is beyond the scope of this paper.

Finally, we want to stress that the importance of viewpoint is significantly decreased in our metric learning approach.
Some classes may exhibit a high degree of freedom such as the plane, but the retrieval results are also excellent (as shown in Fig. \ref{fig:tub}).

\subsubsection{Analysis}


We further show some statistics on this dataset.
First, we provide the precision-recall values at fixed points in Table \ref{tab:pr}.
Compared to Fig. 9 in \cite{eitz2012sbsr}, our results are approximately $10 \%$ higher.
We then show six standard evaluation metrics in Table \ref{tab:psbmetric}.
Since other methods did not report the results on this dataset, we leave the comprehensive comparison to the next section.
Instead, in this analysis we focus on the effectiveness of metric learning for shape retrieval.

PSB/SBSR is a very imbalanced dataset, where training and testing only partially overlap. 
Namely, there are 21 classes appear in both training and testing sets, while 71 classes are used solely for testing.
This makes it an excellent dataset for investigating similarity learning, because the ``unseen'' classes verify the learning is not biased.

\begin{figure}[htbp]
\begin{center}
\begin{tabular}{c}
	\includegraphics[width=0.9\columnwidth]{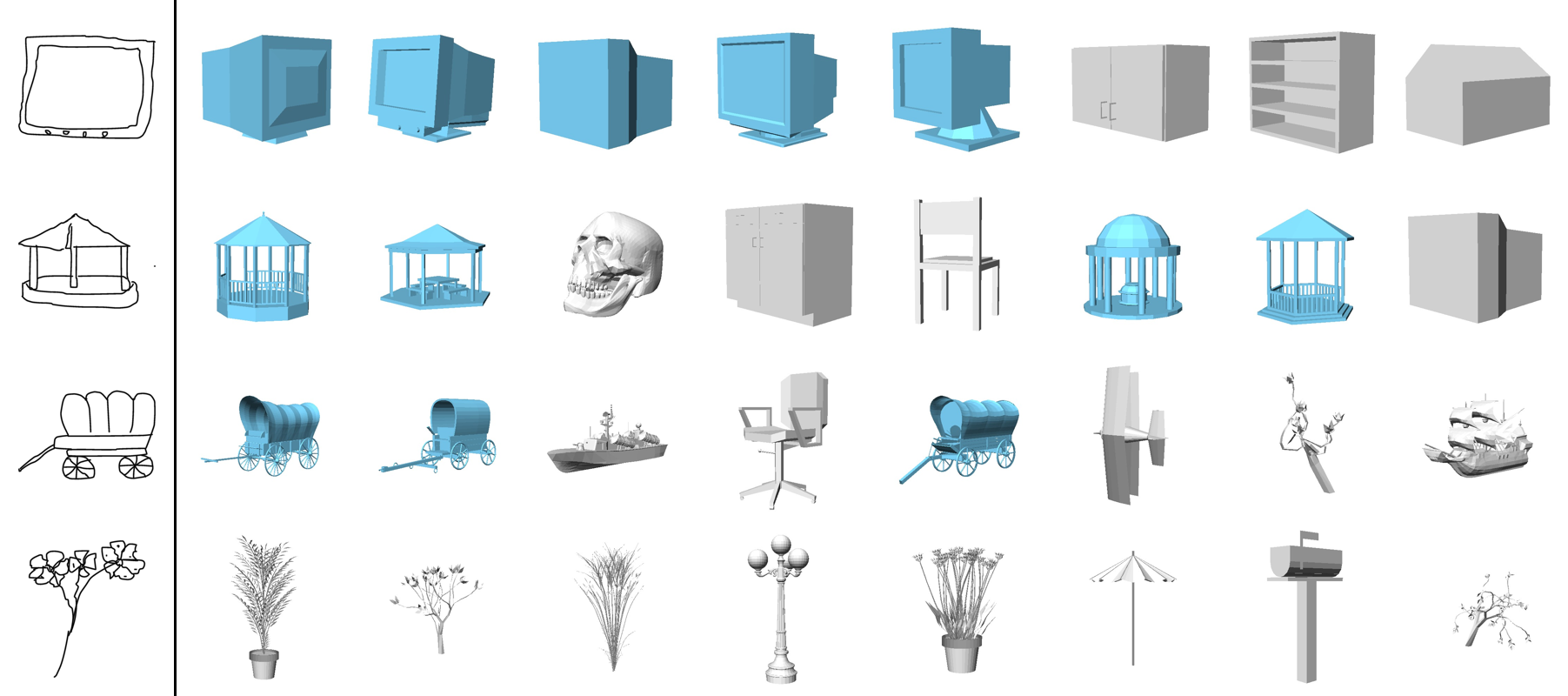}
\end{tabular}
\end{center}
\caption{Retrieval examples of unseen samples in PSB/SBSR dataset. The cyan denotes the correct retrievals.}
\label{fig:tub-test-only}
\end{figure}

We show some examples for these unseen classes in Fig. \ref{fig:tub-test-only} (more statistical curves are available on project website due to the space limitation).
It is interesting to see that our proposed method works well even on failure cases ({\it e.g.,} the flower), where the retrieval returns similar shapes (``potting plant'').
This demonstrates that our method learns the similarity effectively.

\begin{table}[h]
\caption{Precision-recall on fixed points.}
\begin{center}
\begin{tabular}{c|c|c|c|c|c}

   \hline 
    5$\%$ &  20$\%$ & 40$\%$ & 60$\%$ & 80$\%$ & 100$\%$   \\ \hline 
   0.616 & 0.286 & 0.221 & 0.180 & 0.138 & 0.072   \\ \hline
  
\end{tabular}
\end{center}
\label{tab:pr}
\end{table}


\begin{table}[h]
\caption{Standard metrics on the PSB/SBSR dataset.}
\begin{center}
\begin{tabular}{c|c|c|c|c|c}

   \hline 
    NN &  FT & ST & E & DCG  & mAP  \\ \hline
   0.223 & 0.177 & 0.271 & 0.173 & 0.451 & 0.218   \\ \hline
  
\end{tabular}
\end{center}
\label{tab:psbmetric}
\end{table}



\subsection{Shape retrieval on SHREC'13 dataset}
In this section, we use the SHREC'13 benchmark to evaluate our method.
We also show the retrieval results within the same domain. 

\vspace{-5pt}
\subsubsection{A visualization of the learned features}
\begin{figure}[htbp]
\begin{center}
\begin{tabular}{c}
	\includegraphics[width=0.95\columnwidth]{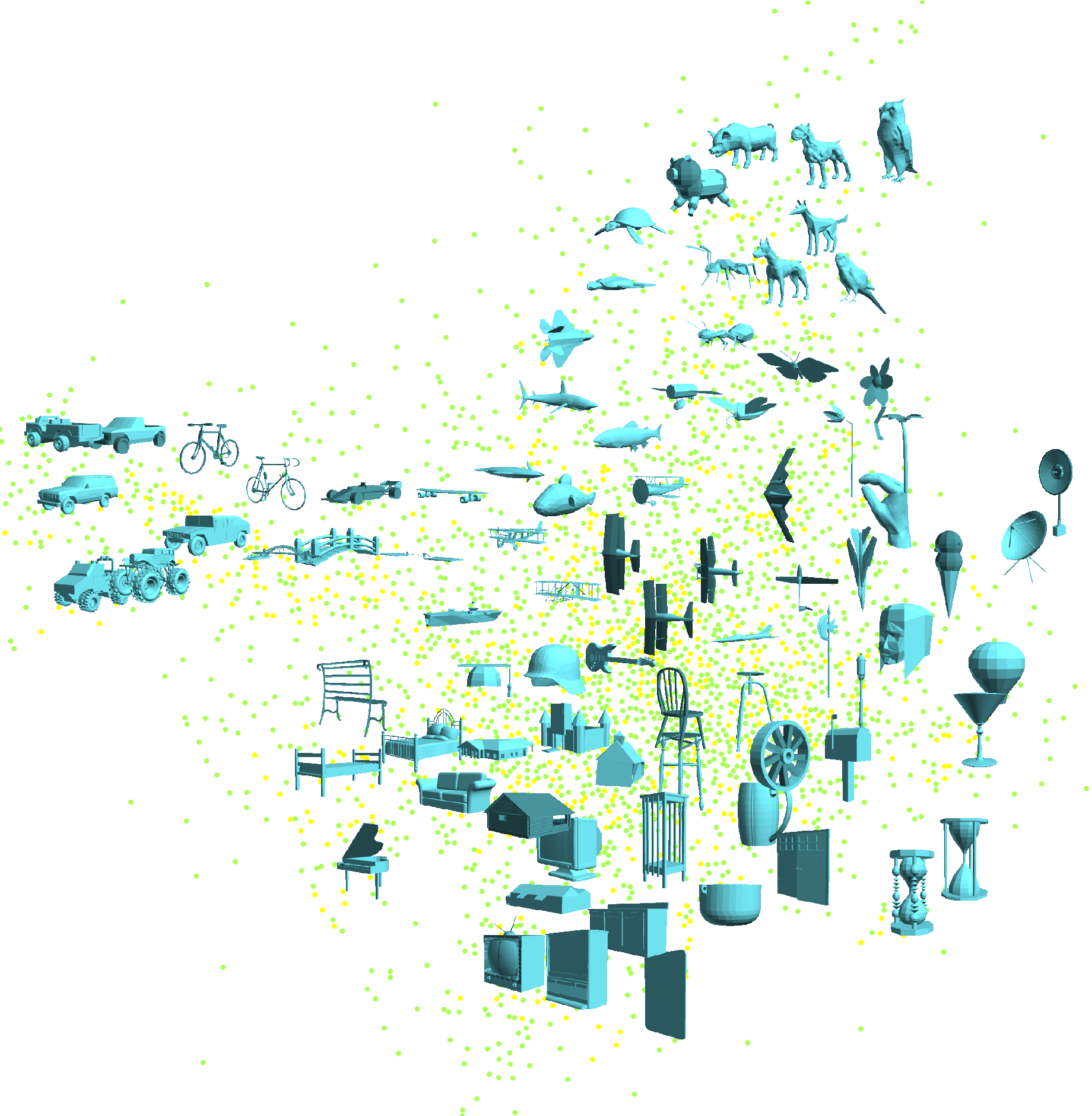}
\end{tabular}
\end{center}
\caption{Visualization of feature space on SHREC'13. Sketch and view feature points are shown by green \& yellow, respectively.}
\label{fig:vis}
\end{figure}

First, we present a visualization of our learned features in Fig. \ref{fig:vis}.
We perform PCA on the learned features and reduce the dimension to two for visualization.
The green dots denote the sketches, and the yellow ones denote views.
For simplicity, we only overlay the views over the point cloud.
Please visit  \url{http://users.cecs.anu.edu.au/~yili/cnnsbsr/} for an interactive demo.

While this is a coarse visualization, we can already see some interesting properties of our method.
First, we can see that classes with similar shapes are grouped together automatically.
On the top right, different animals are mapped to neighboring positions.
On the left, various types of vehicles are grouped autonomously.
Other examples include house and church, which are very similar.
Note that this is an weakly supervised method. This localization suggests that the learned features are very useful for both within-domain and cross domain retrievals.

\vspace{-5pt}
\subsubsection{Statistical results}
We present the statistical results on SHREC'13 in this section.
First, we compare the precision-recall curve against the state of the art methods reported in \cite{li2014comparison}.
\begin{figure*}[htbp]
\begin{center}
\begin{tabular}{cc}
	\includegraphics[height=5cm]{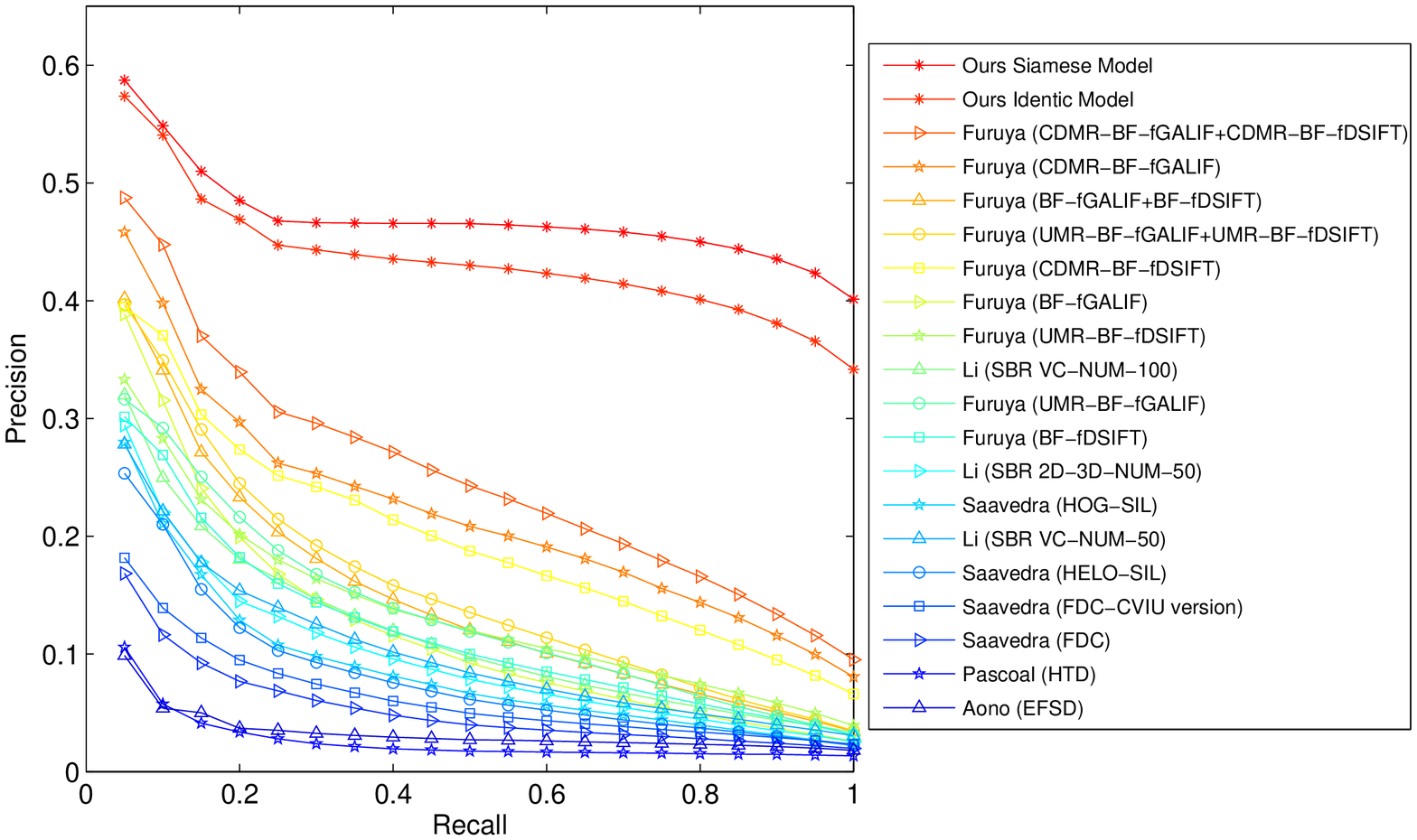}  &
	\includegraphics[height=5cm]{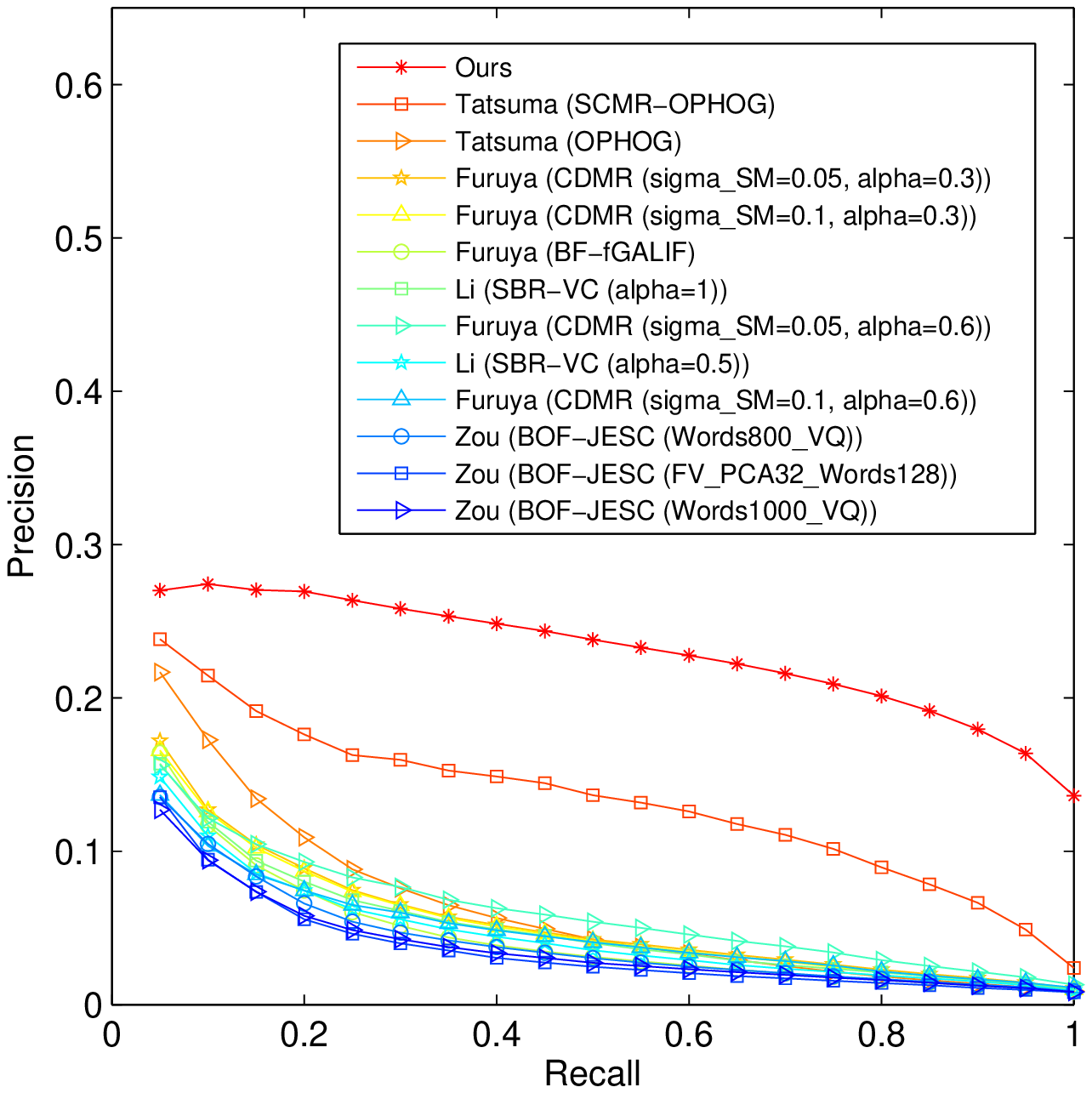} \\
	(a) SHREC'13 comparison  &  (b) SHREC'14 comparison\\
\end{tabular}
\end{center}
\caption{Performance comparison on SHREC'13 \& '14. Please refer to \cite{li2014comparison} and \cite{LLL+14a} for the descriptions of the compared methods.}
\label{fig:SHRECstat}
\end{figure*}

%

From the Fig. \ref{fig:SHRECstat} we can see that our method significantly outperforms other comparison methods.
On SHREC'13 benchmark, the performance gain of our method is already $10 \%$ when recall is small.
More importantly, the whole curve decreases much slower than other methods when the recall increases, which is desirable because it shows the method is more stable.
Our method has a higher performance gain ($30 \%$) when recall reaches $1$.


We note that there is a noticeable overfitting in the training when a stopping criterion is reached.
It suggests the performance can be even better, if one can fine tune and explore the network structure and training procedure.

We further show the standard metrics for comparison.
These metrics examine the retrieval from different perspectives.
For simplicity, we only select the best method from each research group in  \cite{li2014comparison}.
As shown in Table \ref{tab:shreccmp}, our method performs better in every metric on both benchmarks.
This further demonstrates our method is superior.


\begin{table}
\caption{Comparison on SHREC'13 \& '14 dataset. The best results are shown in red, and the second best results are shown in blue.}
\begin{center}
\begin{tabular}{c|c|c|c|c|c|c}

 \hline

\multicolumn{7}{c}{SHREC'13} \\ \hline
 
   &  NN    &   FT    &   ST    &   E   &    DCG  &  mAP \\ \hline \hline     
  
Ours         &   \color{red}{0.405} & \color{red}{0.403} & \color{red}{0.548} & \color{red}{0.287} & \color{red}{0.607}  & \color{red}{0.469}  \\ \hline
Identic   &   \color{blue}{0.389} & \color{blue}{0.364} & \color{blue}{0.516} & \color{blue}{0.272} & \color{blue}{0.588}  & \color{blue}{0.434} \\ \hline \hline

\cite{furuya2013ranking}  & 0.279  & 0.203  & 0.296  & 0.166  & 0.458  & 0.250 \\ \hline   
\cite{li2014comparison}  & 0.164  & 0.097  & 0.149  & 0.085  & 0.348  & 0.116 \\ \hline       
\cite{journals/vlc/SousaF10}  & 0.017  & 0.016  & 0.031  & 0.018  & 0.240  & 0.026 \\ \hline 
\cite{Saavedra:2012:SMR:2381194.2381204}   & 0.110  & 0.069  & 0.107  & 0.061  & 0.307  & 0.086 \\ \hline  

\hline

\multicolumn{7}{c}{SHREC'14} \\ \hline
   &  NN    &   FT    &   ST    &   E   &    DCG  &  mAP \\ \hline \hline     
  
Ours         &   \color{red}{0.239} & \color{red}{0.212} & \color{red}{0.316} & \color{red}{0.140} & \color{red}{0.496}  & \color{red}{0.228}  \\ \hline
\hline
\cite{tatsuma2012large}  & 0.160  & 0.115  & 0.170  & 0.079  & 0.376  & 0.131  \\ \hline  
\cite{furuya2013ranking}  & 0.109  & 0.057  & 0.089  & 0.041  & 0.328  & 0.054  \\ \hline  
\cite{li2014comparison}  & 0.095  & 0.050  & 0.081  & 0.037  & 0.319  & 0.050  \\ \hline 

\end{tabular}
\end{center}
\label{tab:shreccmp}
\end{table}

%
%
%

We also compare to the case where both networks are identical, \textit{i.e.}, both views and sketches use the same Siamese network.
Fig. \ref{fig:SHRECstat}a suggests that this configuration is inferior than our proposed version, but still it is better than all other methods.
This supports our hypothesis that the variations in two domains are different.
This also sends a message that using the same features (hand-crafted or learned) for both domains may not be ideal.

\subsubsection{Within-domain retrieval}
Finally, we show the retrievals in the same domain.
This interesting experiment shall be straightforward to report because the data is readily available, but was not shown before in any literature.
Since this is a ``by-product'' of our method, we do not tune up any parameter or re-train the system.

\begin{figure}[htbp]
\begin{center}
\begin{tabular}{c}
	\includegraphics[width=0.9\columnwidth]{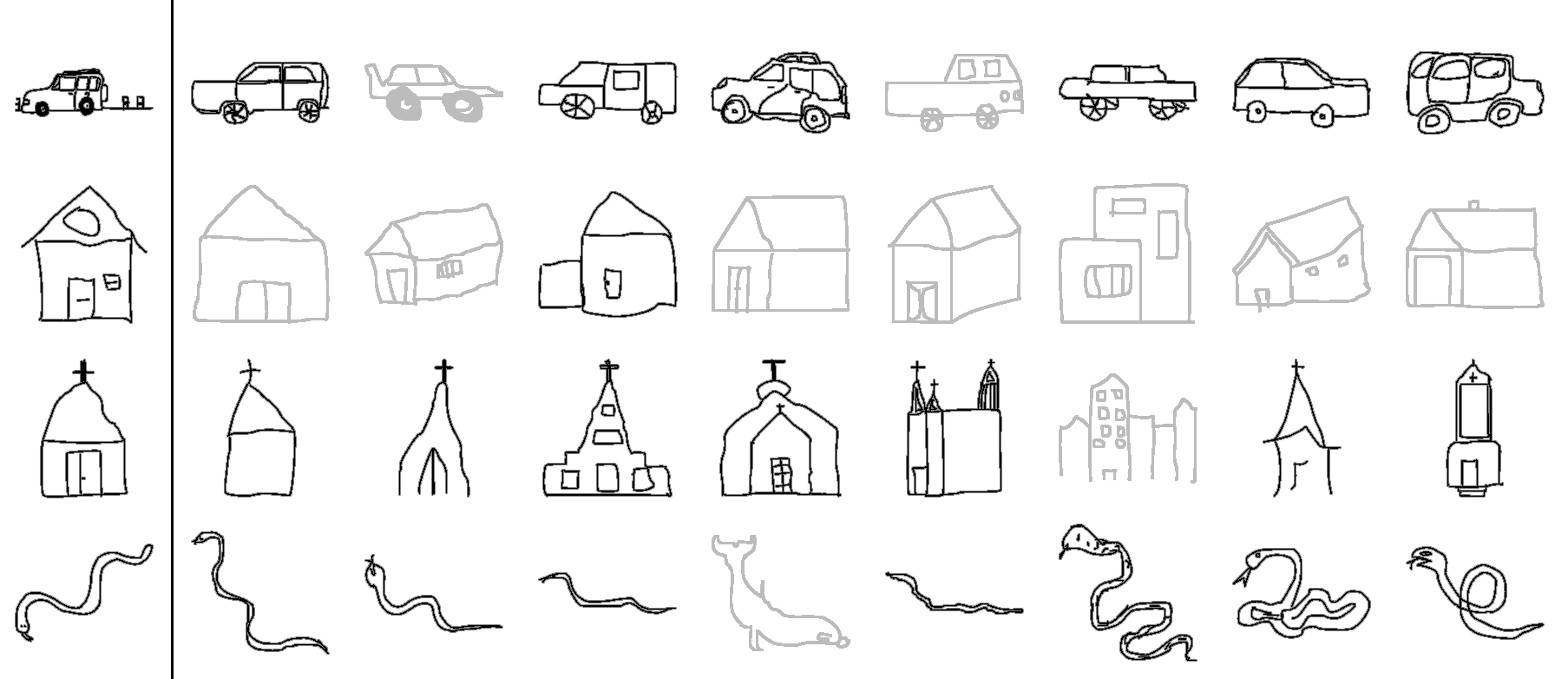} \\
\end{tabular}
\end{center}
\caption{Sketch-sketch retrieval for SHREC'13. The incorrect retrievals are marked as light gray.}
\label{fig:withindomain}
\end{figure}

\begin{figure}[htbp]
\begin{center}
\begin{tabular}{c}
	\includegraphics[width=0.9\columnwidth]{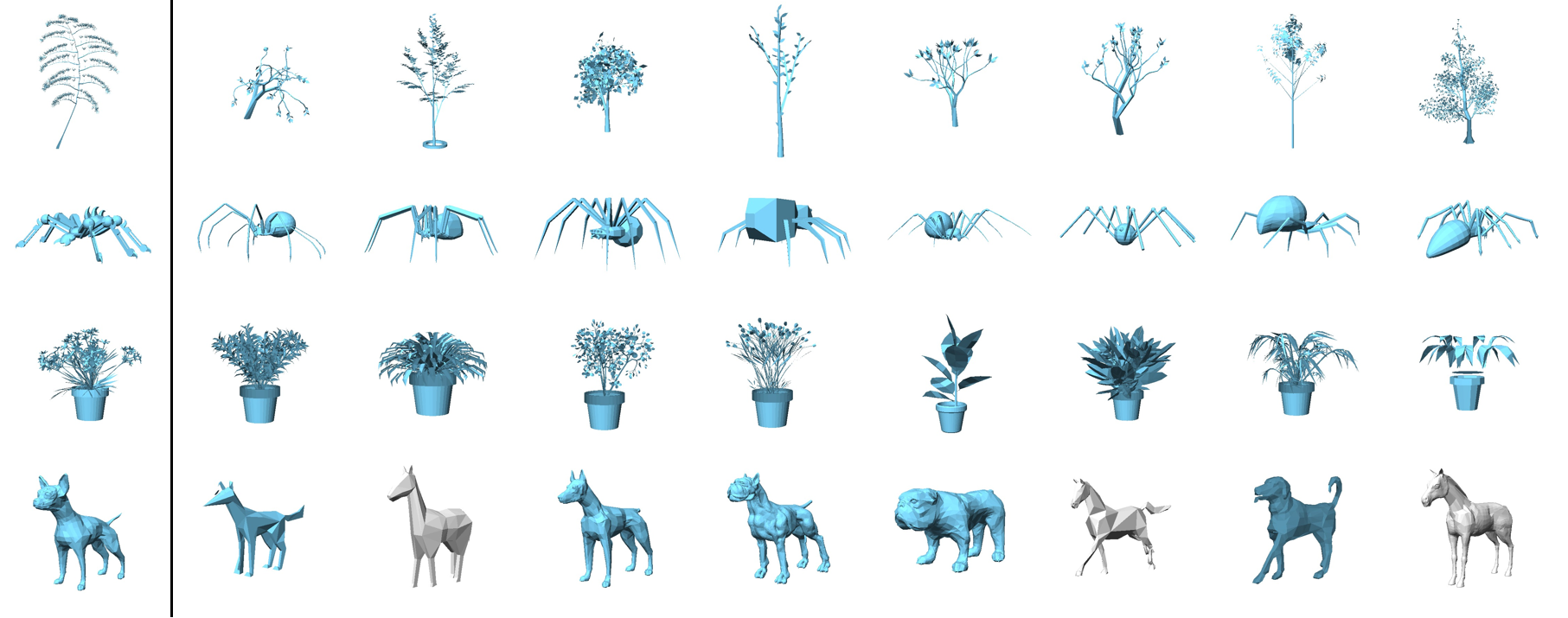}
\end{tabular}
\end{center}
\caption{View-view retrieval for SHREC'13. The cyan denotes the correct retrievals.}
\label{fig:withindomain2}
\end{figure}

Figs. \ref{fig:withindomain} and \ref{fig:withindomain2} visualize some retrieval results in each domain, respectively.
Table \ref{tab:withindomain} further reports the statistics.
The retrieval results demonstrate our method is powerful in learning the features for both within-domain and cross-domain.
From these figures, one can see that the view domain is much more consistent than the sketch domain.
Comparing Table \ref{tab:withindomain} to Table \ref{tab:shreccmp}, we conclude that the inconsistency in sketch is the most challenging issue in the sketch based 3D shape retrieval.

\begin{table}
\caption{Standard metrics for the within-domain retrieval on SHREC'13.}
\begin{center}
\begin{tabular}{c|c|c|c|c|c|c}

   \hline 
             &   NN &  FT & ST & E & DCG  & mAP  \\ \hline 
 view  &   0.965 & 0.877 & 0.982 & 0.536 & 0.971 & 0.909   \\ \hline
 sketch  & 0.431 & 0.352 & 0.514 & 0.298 & 0.679 & 0.373   \\ \hline
 
\end{tabular}
\end{center}
\label{tab:withindomain}
\end{table}

\section{Conclusion}
In this paper we propose to learn feature presentations for sketch based 3D shape retrieval. Instead of computing ``best views'' and match them against queries, we use predefined viewpoints for the whole dataset and adopt two Siamese CNNs, one for views and one for sketches.
Our experiments on three large datasets demonstrated that our method is superior. 

\section{Acknowledgement}
NICTA is funded by the Australian Government
as represented by the Department of Broadband, Communications and the Digital Economy and the Australian Research Council through the ICT Centre of Excellence program.

{\small \bibliographystyle{ieee}
\bibliography{CVPR-2015-3dShapeRetrieval}

\begin{thebibliography}{10}\itemsep=-1pt

\bibitem{belongie2002shape}
S.~Belongie, J.~Malik, and J.~Puzicha.
\newblock Shape matching and object recognition using shape contexts.
\newblock {\em IEEE Trans. Pattern Anal. Mach. Intell.}, 24(4):509--522, Apr.
  2002.

\bibitem{Theano}
J.~Bergstra, O.~Breuleux, F.~Bastien, P.~Lamblin, R.~Pascanu, G.~Desjardins,
  J.~Turian, D.~Warde-Farley, and Y.~Bengio.
\newblock Theano: a {CPU} and {GPU} math expression compiler.
\newblock In {\em Proceedings of the Python for Scientific Computing Conference
  ({SciPy})}, Jun. 2010.

\bibitem{Bronstein:2011:SGG:1899404.1899405}
A.~M. Bronstein, M.~M. Bronstein, L.~J. Guibas, and M.~Ovsjanikov.
\newblock Shape google: Geometric words and expressions for invariant shape
  retrieval.
\newblock {\em ACM Trans. Graph.}, 30(1):1:1--1:20, Feb. 2011.

\bibitem{NIPS2011_4314}
K.~Chen and A.~Salman.
\newblock Extracting speaker-specific information with a regularized siamese
  deep network.
\newblock In J.~Shawe-Taylor, R.~Zemel, P.~Bartlett, F.~Pereira, and
  K.~Weinberger, editors, {\em NIPS 2011}, pages 298--306. 2011.

\bibitem{chopra2005learning}
S.~Chopra, R.~Hadsell, and Y.~LeCun.
\newblock Learning a similarity metric discriminatively, with application to
  face verification.
\newblock In {\em CVPR 2005}, volume~1, pages 539--546. IEEE, 2005.

\bibitem{journals/ijcv/DarasA10}
P.~Daras and A.~Axenopoulos.
\newblock A 3d shape retrieval framework supporting multimodal queries.
\newblock {\em International Journal of Computer Vision}, 89(2-3):229--247,
  2010.

\bibitem{DeCarlo:2003:SCF}
D.~DeCarlo, A.~Finkelstein, S.~Rusinkiewicz, and A.~Santella.
\newblock Suggestive contours for conveying shape.
\newblock {\em ACM Trans on Graphics}, 22(3):848--855, July 2003.

\bibitem{eitz2012hdhso}
M.~Eitz, J.~Hays, and M.~Alexa.
\newblock How do humans sketch objects?
\newblock {\em ACM Trans. on Graphics}, 31(4):44:1--44:10, 2012.

\bibitem{eitz2012sbsr}
M.~Eitz, R.~Richter, T.~Boubekeur, K.~Hildebrand, and M.~Alexa.
\newblock Sketch-based shape retrieval.
\newblock {\em ACM Trans. Graphics}, 31(4):31:1--31:10, 2012.

\bibitem{Funkhouser:2003:ASE}
T.~Funkhouser, P.~Min, M.~Kazhdan, J.~Chen, A.~Halderman, D.~Dobkin, and
  D.~Jacobs.
\newblock A search engine for {3D} models.
\newblock {\em ACM Transactions on Graphics}, 22(1):83--105, Jan. 2003.

\bibitem{Funkhouser:2006:PMS:1281957.1281974}
T.~Funkhouser and P.~Shilane.
\newblock Partial matching of 3d shapes with priority-driven search.
\newblock In {\em Proceedings of the Fourth Eurographics Symposium on Geometry
  Processing}, pages 131--142. Eurographics Association, 2006.

\bibitem{furuya2009dense}
T.~Furuya and R.~Ohbuchi.
\newblock Dense sampling and fast encoding for 3d model retrieval using
  bag-of-visual features.
\newblock In {\em Proceedings of the ACM international conference on image and
  video retrieval}, page~26. ACM, 2009.

\bibitem{furuya2013ranking}
T.~Furuya and R.~Ohbuchi.
\newblock Ranking on cross-domain manifold for sketch-based 3d model retrieval.
\newblock In {\em International Conference on Cyberworlds 2013}, pages
  274--281. IEEE, 2013.

\bibitem{gong2013learning}
B.~Gong, J.~Liu, X.~Wang, and X.~Tang.
\newblock Learning semantic signatures for 3d object retrieval.
\newblock {\em Multimedia, IEEE Transactions on}, 15(2):369--377, 2013.

\bibitem{Kazhdan:2002:RSD:645316.649205}
M.~M. Kazhdan, B.~Chazelle, D.~P. Dobkin, A.~Finkelstein, and T.~A. Funkhouser.
\newblock A reflective symmetry descriptor.
\newblock ECCV '02, pages 642--656, London, UK, UK, 2002. Springer-Verlag.

\bibitem{Krizhevsky2012}
A.~Krizhevsky, I.~Sutskever, and G.~Hinton.
\newblock Imagenet classification with deep convolutional neural networks.
\newblock In {\em NIPS}, 2012.

\bibitem{LeCun:1998:CNI:303568.303704}
Y.~LeCun and Y.~Bengio.
\newblock The handbook of brain theory and neural networks.
\newblock chapter Convolutional networks for images, speech, and time series,
  pages 255--258. MIT Press, Cambridge, MA, USA, 1998.

\bibitem{li2014comparison}
B.~Li, Y.~Lu, A.~Godil, T.~Schreck, B.~Bustos, A.~Ferreira, T.~Furuya, M.~J.
  Fonseca, H.~Johan, T.~Matsuda, et~al.
\newblock A comparison of methods for sketch-based 3d shape retrieval.
\newblock {\em Computer Vision and Image Understanding}, 119:57--80, 2014.

\bibitem{li2015comparison}
B.~Li, Y.~Lu, C.~Li, A.~Godil, T.~Schreck, M.~Aono, M.~Burtscher, Q.~Chen,
  N.~K. Chowdhury, B.~Fang, et~al.
\newblock A comparison of 3d shape retrieval methods based on a large-scale
  benchmark supporting multimodal queries.
\newblock {\em Computer Vision and Image Understanding}, 131:1--27, 2015.

\bibitem{LLL+14a}
B.~Li, Y.~Lu, C.~Li, A.~Godil, T.~Schreck, M.~Aono, Q.~Chen, N.~Chowdhury,
  B.~Fang, T.~Furuya, H.~Johan, R.~Kosaka, H.~Koyanagi, R.~Ohbuchi, and
  A.~Tatsuma.
\newblock {SHREC’14 track: Comprehensive 3d shape retrieval}.
\newblock In {\em Proc. EG Workshop on 3D Object Retrieval}, 2014.

\bibitem{Osada:2002:SD}
R.~Osada, T.~Funkhouser, B.~Chazelle, and D.~Dobkin.
\newblock Shape distributions.
\newblock {\em ACM Transactions on Graphics}, 21(4):807--832, Oct. 2002.

\bibitem{Saavedra:2012:SMR:2381194.2381204}
J.~M. Saavedra, B.~Bustos, T.~Schreck, S.~Yoon, and M.~Scherer.
\newblock Sketch-based 3d model retrieval using keyshapes for global and local
  representation.
\newblock In {\em EG 3DOR'12}, pages 47--50. Eurographics Association, 2012.

\bibitem{shilane2004princeton}
P.~Shilane, P.~Min, M.~Kazhdan, and T.~Funkhouser.
\newblock The princeton shape benchmark.
\newblock In {\em Shape modeling applications, 2004. Proceedings}, pages
  167--178. IEEE, 2004.

\bibitem{journals/vlc/SousaF10}
P.~M.~A. Sousa and M.~J. Fonseca.
\newblock Sketch-based retrieval of drawings using spatial proximity.
\newblock {\em J. Vis. Lang. Comput.}, 21(2):69--80, 2010.

\bibitem{Tangelder:2008:SCB:1395016.1395041}
J.~W. Tangelder and R.~C. Veltkamp.
\newblock A survey of content based 3d shape retrieval methods.
\newblock {\em Multimedia Tools Appl.}, 39(3):441--471, Sept. 2008.

\bibitem{tatsuma2012large}
A.~Tatsuma, H.~Koyanagi, and M.~Aono.
\newblock A large-scale shape benchmark for 3d object retrieval: Toyohashi
  shape benchmark.
\newblock {\em Signal and Information Processing Association}, pages 3--6,
  2012.

\bibitem{Wessel:2009:SBR:2381128.2381141}
R.~Wessel, I.~Bl\"{u}mel, and R.~Klein.
\newblock A 3d shape benchmark for retrieval and automatic classification of
  architectural data.
\newblock In {\em 3DOR '09}, pages 53--56. Eurographics Association, 2009.

\bibitem{Yih:2011:LDP:2018936.2018965}
W.~Yih, K.~Toutanova, J.~C. Platt, and C.~Meek.
\newblock Learning discriminative projections for text similarity measures.
\newblock In {\em Proceedings of the Fifteenth Conference on Computational
  Natural Language Learning}, CoNLL '11, pages 247--256, 2011.

\end{thebibliography}
 }
\end{document}